%% file: paper.tex
\documentclass[11pt]{article}

\usepackage{times}
\usepackage{url}
\usepackage{amsmath}
\usepackage{graphicx}
\usepackage{natbib}
\usepackage{comment}
\usepackage{algorithm}
\usepackage{algpseudocode}
\usepackage{rotating}
\usepackage{lscape}
\usepackage{xcolor,colortbl}
\usepackage{hyperref}
\usepackage{multirow}
\usepackage{pgfplots}
\usepackage{placeins}

\input macros

\begin{document}
\pagenumbering{arabic}

\title{A Brief Summary of Explanatory Virtues}

\author{Ingrid Zukerman\\
Department of Data Science and Artificial Intelligence\\
        Faculty of Information Technology\\
	Monash University, Clayton, Victoria 3800, Australia\\
        {\ttsm ingrid.zukerman{@}monash.edu}}    

\date{}
\maketitle
\begin{abstract}
In this report, I provide a brief summary of the literature in philosophy,
psychology and cognitive science about Explanatory Virtues, and link
these concepts to eXplainable AI.
\end{abstract}

\section{Introduction}
\textit{Explanatory Virtues} (\textit{EVs}\/) have been mainly
discussed in the context of abductive reasoning (inference to the
best explanation), where the explanation is a theory that accounts
for observations. Over the years, practitioners in philosophy,
psychology and cognitive science have studied various EVs, and coined
many terms for them. In this report, I largely follow the taxonomy
in~\citep{keas2018systematizing}, which comprises four main types of
EVs: \textit{evidential}, \textit{coherential}, \textit{aesthetic}
and \textit{diachronic} (Section~\ref{section:terms}). I have included
additional EVs to those considered by \cite{keas2018systematizing},
and added one category, viz \textit{coverage}, that pertains to the
wider applicability of a theory, and accounts for EVs that don't fit
well into \citeauthor{keas2018systematizing}'s taxonomy.

It is worth noting that in the context of \textit{eXplainable AI}
(\textit{XAI}\/), for \textit{global explanations}, which explain
an entire model, \eg~\citep{BastaniEtAl2017,LakkarajuEtAl2017}, the
``observations'' are the way a Machine Learning (ML) model works and
its parameters, and the ``theories'' are explanations that account for
these observations. For \textit{local explanations}, which pertain
to the outcome predicted by an ML model for a specific instance,
\eg~\citep{Ribeiro:16,biran2017human}, the observations also include
the feature values of the instance in question and the prediction made
by the model for this instance.

The main EV dimensions are described in Section~\ref{section:dimensions},
followed by EVs categories according to our expanded taxonomy, in
Section~\ref{section:terms}. In Section~\ref{section:EVs-XAI}, I discuss
how EVs apply to XAI.

\section{Main EV dimensions}
\label{section:dimensions}
There is a large body of literature about EVs in philosophy, psychology
and cognitive science. Here, I provide categorizations of these EVs along
two dimensions: \textit{epistemic/pragmatic}~\citep{vanFraassen1980}
and \textit{ontological}~\citep{keas2018systematizing}.

\myparagraph{\bem Epistemic/pragmatic dimension.}
\cite{vanFraassen1980} distinguishes between two types of EVs according
to their roles: \textit{epistemic} virtues, which mean that explanations
(theories) are valuable in themselves as a means to understanding the
world, and \textit{pragmatic} virtues, which are concerned with the
use and usefulness of a theory, and provide reasons to prefer a theory
independently of questions of truth.\footnote{\cite{lipton2004inference}
and \cite{glymour2015probability} also make such a distinction, but their
account is less complete than the account in \citep{vanFraassen1980}.}

\begin{itemize}
\item 
\textit{Epistemic virtues} -- Consistency (internally and with the facts),
Empirical adequacy (how well a model's predictions match the data) and
Empirical strength (a model is empirically stronger than a rival if its
theoretical constructs are subsumed by those required by the rival --
similar to Modesty~\citep{QuineUllian1978}).

\item
\textit{Pragmatic virtues} -- Mathematical elegance, Simplicity (a model has
few theoretical requirements), Completeness (a model explains most of
the data), Unification (a model accounts for additional phenomena)
and Explanatory power (a model decreases the surprisingness of an
observation).
\end{itemize}

\myparagraph{\bem Ontological dimension.}
The categories in this dimension are largely based on the ontology
in~\citep{keas2018systematizing}, which comprises four main types
of EVs: evidential, coherential, aesthetic and diachronic. As
mentioned above, I have included additional EVs from the literature in
\citeauthor{keas2018systematizing}'s (\citeyear{keas2018systematizing})
categories, and added one category, viz coverage, to account for
EVs that do not fit well into \citeauthor{keas2018systematizing}'s
ontology.\footnote{Unification, which I deem to be a coverage virtue,
was categorized by \cite{keas2018systematizing} as an aesthetic virtue.}

\section{EVs in the ontological dimension}
\label{section:terms}
Here, I present definitions of EVs from the literature, and incorporate
them in the categorization of \cite{keas2018systematizing}.
Table~\ref{tab:evlitauthor} lists these EVs by author, and
Table~\ref{tab:evlit} lists them according to their category, and includes
terms used by different authors for similar EVs.

\myparagraph{Evidential virtues.} 
These virtues indicate different facets of how well a theory accounts for observations in the world.
\begin{itemize}
\myitem 
\textit{Evidential
accuracy}~\citep{Kuhn1977,vanFraassen1980,thagard1989coherence,lipton1991inference,RosalesMorton2021}
-- A theory $T$ fits the empirical evidence well (regardless of causal
claims).

\item
\textit{Causal adequacy}~\citep{thagard1989coherence,lipton1991inference,Psillos2002,mackonis2013inference,RosalesMorton2021}
-- A theory $T$'s causal factors plausibly produce the effects (evidence)
in need of explanation.

\item
\textit{Refutability}~\citep{QuineUllian1978,vancleave2016} --
Some imaginable event must suffice to refute the hypothesis/theory $T$.

\item
\textit{Explanatory depth}~\citep{HitchcockWoodward2003} -- A theory $T$
excels in causal history depth (how far back a causal chain goes) or in
other depth measures, such as the range of counterfactual questions that
theory $T$ answers regarding the item being explained.

\item
\textit{Optimality}~\citep{McMullin2014} -- Whether a theory $T$ affords
the best explanation available (inference to the best explanation).
\end{itemize}

\input tables/ev-lit-tab-author

\input tables/ev-lit-tab

\myparagraph{Coherential virtues.}
These virtues describe how well the components of a theory agree with
each other and with other warranted beliefs.
\begin{itemize}
\item
\textit{Internal coherence}~\citep{McMullin2014} -- a theory $T$ lacks
\adhoc\ hypotheses -- theoretical components merely tacked on to solve
isolated problems.

\item
\textit{Internal
consistency}~\citep{Kuhn1977,vanFraassen1980,Newton-Smith1981,thagard1989coherence,McMullin2014}
-- A theory $T$'s components should not be contradictory and should
be coordinated into an intuitively plausible whole.

\item
\textit{Universal
consistency}~\citep{hempel1948studies,Ausubel1962,Kuhn1977,QuineUllian1978,vanFraassen1980,Newton-Smith1981,Psillos2002,mackonis2013inference,McMullin2014}
-- A theory $T$ sits well with (or is not obviously contrary to) other
warranted beliefs.

\item
\textit{Analogy}~\citep{thagard1978best,thagard1989coherence} relies on
the similarity between an explanandum and a familiar, well-established
causal mechanism. Analogy does not guarantee that a theory is correct,
but increases its value.

\item
\textit{Compatibility with metaphysical beliefs}~\citep{Newton-Smith1981}
-- A theory $T$ agrees with established practices in the field.

\item
\textit{Conservativeness}~\citep{vanFraassen1980,schupbach2011logic,vancleave2016}
-- A theory $T$'s ability to decrease the degree to which we
find the explanandum surprising (increase in expectedness);
it should force us to give up fewer well established
beliefs~\citep{vancleave2016}.

\item
\textit{Depth}~\citep{vancleave2016,lipton1991inference} -- A theory
should not posit fundamentally new types of phenomena (raise more
questions than it answers).
\end{itemize}

\myparagraph{Coverage virtues.}
These virtues pertain to the applicability of a theory beyond the observations
for which it was developed.
\begin{itemize}
\item 
\textit{Completeness}~\citep{Psillos2002,thagard1989coherence,vancleave2016}
-- A theory $T$ explains more observations than rival explanations.

\item
\textit{Scope}~\citep{Kuhn1977,vanFraassen1980,QuineUllian1978} -- The
range of application of a theory $T$ should extend beyond the particular
observations it was initially designed to explain.

\item
\textit{Consilience}~\citep{thagard1978best} --
A theory $T$ should explain the target observations, observations in
more than one domain, and should specify what it cannot explain.

\item
\textit{Unification}~\citep{vanFraassen1980,lipton1991inference,Psillos2002,mackonis2013inference,RosalesMorton2021,vancleave2016}
-- A theory $T$ explains more kinds of facts than rival explanations with
the same amount of theoretical content.

\item
\textit{Importance}~\citep{Psillos2002} -- A theory $T$ is more important
than a rival explanation if the phenomena explained by $T$ are more
salient than those explained by the rival.
\end{itemize}

Completeness poses no constraints on the number of theoretical
requirements of an explanation. Hence, Completeness is entailed by
Unification. Since Unification poses no requirements regarding what a
theory cannot explain, it is entailed by Consilience.

\myparagraph{Aesthetic virtues.}
These virtues consider different aspects of brevity as desirable
features of an explanation.
\begin{itemize}
\item
\textit{Simplicity}~\citep{Kuhn1977,thagard1978best,QuineUllian1978,vanFraassen1980,lipton1991inference,Psillos2002,mackonis2013inference,vancleave2016}
 -- $T$ explains the same facts as rival explanations, but with fewer
theoretical requirements. 

\item
\textit{Modesty}~\citep{QuineUllian1978,vanFraassen1980,vancleave2016}
-- A theory $T$ is more modest than another if it is weaker in a logical
sense: if it is implied by the other.
\end{itemize}

Both Simplicity and Modesty refer to theoretical requirements (\ie\
concepts or rules mentioned in an explanation). Since Modesty demands
a subset relation between the requirements of rival explanations, it
implies Simplicity.

\myparagraph{Diachronic virtues.}
These virtues are concerned with the way a theory is affected by time. 
\begin{itemize}
\item 
\textit{Durability}~\citep{Newton-Smith1981,McMullin2014} -- A theory $T$ has survived
testing by successfully predicting or plausibly accommodating new data.

\item
\textit{Fruitfulness}~\citep{Kuhn1977,Newton-Smith1981,McMullin2014}
-- A theory $T$ has generated additional discovery by means such as
successful novel prediction, unification, and non \adhoc\ theoretical
elaboration.

\item
\textit{Smoothness}~\citep{Newton-Smith1981} -- The smoothness with
which adjustments can be made to a theory $T$ in the face of failure.
\end{itemize}

As mentioned above, \cite{vanFraassen1980} distinguishes between
\textit{epistemic} and \textit{pragmatic} virtues. The former means
that theories (in our case, explanations) are valuable in themselves
as a way of understanding the world, and the latter provide reasons
to prefer a theory independently of questions of truth. When matching
\citeauthor{keas2018systematizing}'s (\citeyear{keas2018systematizing})
ontology with \citeauthor{vanFraassen1980}'s (\citeyear{vanFraassen1980})
virtues, evidential and diachronic virtues are epistemic, coverage and
aesthetic virtues are pragmatic, two coherential virtues are epistemic
(Internal and Universal consistency) and four are pragmatic (Internal
coherence, Analogy, Conservativeness and Compatibility with metaphysical
beliefs).

Experimental studies about the impact of EVs on people's perceptions of
explanations are described in~\citep{zemla2017evaluating} and citations
therein. These studies consider operationalized attributes, such as
word count.

\section{EVs in XAI}
\label{section:EVs-XAI}
Here, we focus on local explanations of ML models, and discuss the EVs
that are applicable to explanations as theories, where the observations
are ML models applied to a particular instance.

\myparagraph{Evidential virtues.} 
\begin{itemize}
\item 
\textit{Evidential accuracy} -- The explanation must justify how the ML
model reaches the predicted outcome with respect to the instance. For
example, for regression and logistic regression, this means mentioning
background information and important features; and for decision
trees, this means mentioning background information, every feature in
the path to a prediction and its effect on the probability of a class.

\item
\textit{Causal adequacy} -- This EV is not relevant according to
our conceptualization, as there is no causal effect between theory
(explanation) and observation (our explanations don't cause the ML models
or their outcomes).

\item
\textit{Refutability} -- Since a local explanation pertains to the outcome
of a particular model, the notion of an imaginable instance that
refutes this explanation is not relevant. More generally, there could be
ML models and outcomes for particular instances that refute the mechanism
employed to generate explanations.

\item
\textit{Explanatory depth} -- As for causal accuracy, the causal aspect of
this EV is not relevant to XAI, but the ability of an explanation to handle
counterfactuals and hypotheticals is relevant.

\item
\textit{Optimality} -- The optimality of an explanation may be determined
by a multi-attribute function that calculates the extent to which the
explanation satisfies certain criteria, such as the Gricean maxims of
cooperative conversation~\citep{grice1975logic} --- the importance of each
criterion may be empirically or subjectively determined.
\end{itemize}

\myparagraph{Coherential virtues.}
\begin{itemize}
\item
\textit{Internal coherence} -- Explanations should not posit \adhoc\
facts that were added just to fit a particular outcome.

\item
\textit{Internal consistency} (Gricean maxim of Manner -- clarity
and order~\citep{grice1975logic}) -- Explanations must not have
contradictory components, and must be organized into an intuitively
plausible whole. It is worth noting that an explanation may have a
contradictory component that points out how the outcome of an ML model
seems incorrect or inconsistent with some features of the instance at hand
(Consilience). Such an explanation is still internally consistent.

\item
\textit{Universal consistency} (Gricean maxim of
Quality~\citep{grice1975logic}) --  Explanations must be Universally
consistent in the sense that they must fit (or not be contrary to)
warranted beliefs. For regression, logistic regression and decision
trees, the warranted beliefs are shared knowledge in the form of
background information (\eg\ probabilities obtained from data)
or generally accepted beliefs (\eg\ being obese increases the chance
of a heart attack, or high maintenance cost of a car is bad). As for
internal consistency, an explanation that describes how the outcome of
an ML model contradicts accepted knowledge is still universally coherent.

\item
\textit{Analogy} -- Analogies and similes are good explanatory tools,
if they are sound and intuitive, However, the automatic generation of
analogies remains an open research problem~\citep{HeEtAl2024analogy}.

\item
\textit{Compatibility with metaphysical beliefs} -- Like refutability, this EV
pertains to the approach for generating explanations, which must be in
line with established practices, rather than to a particular explanation
about the outcome of an ML model when applied to an instance.

\item
\textit{Conservativeness} -- As mention above, a conservative explanation
reduces the surprisingness (increases the plausibility) of an outcome. For
regression, logistic regression and decision trees, this would involve
explaining why a feature and/or an outcome that appear compelling are
not predicted, \eg\ the feature in question has no effect or other
features carry more weight.

\item
\textit{Depth}~(Gricean maxim of Quantity~\citep{grice1975logic}) --
Explanations should not include more information than necessary.
\end{itemize}

\myparagraph{Coverage virtues.}
The coverage of an explanation of an ML model measures whether the
explanation describes the extent of the applicability of the model,
\eg\ to other instances and/or types of instances.\footnote{Note that the
applicability of an ML model is different from the ground truth for
that model, as an ML model may produce incorrect results for some of
the instances to which it applies.} 
\begin{itemize}
 \item
\textit{Completeness} -- An explanation $E1$ is more Complete than an
explanation $E2$ if it explains more observations, \ie\ the application
of an ML model to more instances.

\item
\textit{Scope} -- An explanation $E1$ has more Scope than an explanation
$E2$ if it mentions additional types of situations or domains to which
an ML model is applicable.

\item
\textit{Consilience} -- An explanation $E1$ is more Consilient than
an explanation $E2$ if it explains more observations, in more than
one domain, and specifies what it cannot explain.

\item
\textit{Unification} -- An explanation $E1$ is more Unifying than an
explanation $E2$ if it relies on the same amount of theoretical content to
explain more observations. For ML models, ``the same amount on theoretical
content'' can be taken to mean ``the same feature values''. For instance,
the following Unifying Explanation for a logistic regression model that
predicts the acceptability of a car adds an applicability component
(in italics) to an initial feature attribution component: ``this car is
acceptable because it has four doors and a large boot. \textit{Indeed,
70 of 80 cars with four doors and a large boot are deemed acceptable by
the model}''~\citep{MarufEtAl2024}.

\item
\textit{Importance} -- This EV requires that the phenomena explained by an
explanation $E1$ (applications of an ML model to an instance to yield
an outcome) are more salient than those explained by an explanation $E2$.
\end{itemize}

\cite{MarufEtAl2024} showed that Unifying explanations for a logistic
regression model are deemed largely equivalent to Conservative
explanations in terms of how much an explanation is liked, and four
explanatory attributes~\citep{hoffman2018metrics}: completeness,
helpfulness for understanding an AI, absence of extraneous
information, and enticement to act on a prediction. However,
Unifying explanations (and other forms of coverage) have been
studied only by \cite{BucincaEtAl2021} and \cite{MarufEtAl2024},
while Conservative explanations have strong support in the XAI
literature~\citep{biran2017human,guidotti2019black,MarufEtAl2023,
miller2019explanation,Sokol:2020interactive,Stepin:2020,vanderWaa:2018}.
The finding in~\citep{MarufEtAl2024} indicates that explanations embodying
coverage virtues warrant further investigation.

\myparagraph{Aesthetic virtues.}
\begin{itemize}
\item 
\textit{Simplicity} -- An explanation $E1$ is Simpler than an explanation
$E2$ if it has fewer requirements.

\item
\textit{Modesty} -- An explanation $E1$ is more Modest than an explanation
$E2$ if the requirements of $E1$ are a subset of the requirements of $E2$,
\end{itemize}

\myparagraph{Diachronic virtues.}
These virtues pertain to how theories fare over time, and are not applicable
to explanations of ML models. As above, they may apply to the procedures
employed to generate explanations.

\section*{Acknowledgments}
This research was supported in part by grant DP190100006 from the
Australian Research Council.

\bibliography{custom}
\bibliographystyle{apalike}

\end{document}

%% file: macros.tex
\definecolor{cerulean}{rgb}{0.0,0.48,0.65}
\definecolor{caribbean}{rgb}{0.0,0.8,0.6}
\definecolor{paleblue}{rgb}{0.82,0.93,1}
\definecolor{auburn}{rgb}{0.43,0.21,0.1}
\definecolor{byzantium}{rgb}{0.44,0.16,0.39}
\definecolor{byzantine}{rgb}{0.74,0.2,0.64}
\definecolor{violet}{rgb}{0.54,0.17,0.89}
\definecolor{cornell}{rgb}{0.7,0.11,0.11}
\definecolor{lgray}{gray}{0.85}
\definecolor{mgray}{gray}{0.7}
\definecolor{dgreen}{rgb}{0,0.55,0}
\definecolor{mgreen}{rgb}{0,0.7,0}
\definecolor{palered}{rgb}{1,0.70,0.75}
\definecolor{pastelyellow}{rgb}{0.99,0.99,0.65}
\definecolor{purple}{rgb}{0.63,0.13,0.94}

\algnewcommand{\LineComment}[1]{\State \(\triangleright\) #1}  

\newcommand{\eg}{e.g.,}
\newcommand{\ie}{i.e.,}

\newcommand{\adhoc}{{\em ad hoc}}

\newcommand{\bem}{\bfseries \itshape }

\newcommand{\myitem}{\vspace*{-2mm}\item}

\newcommand{\myparagraph}{\vspace*{-3.5mm}\paragraph}

\newcommand{\ttsm}{\tt \small}

%% file: tables/ev-lit-tab-author.tex
\setlength{\tabcolsep}{4pt}
\begin{table}[t]
    \centering
\renewcommand{\baselinestretch}{0.95}\selectfont
\caption{Summary of Explanatory Virtues from psychology, philosophy
and cognitive science by author, in chronological order. If a term is
associated with a term in brackets (with similar semantics), the latter
heads a row in Table~\ref{tab:evlit}; otherwise, the original term
heads a row in Table~\ref{tab:evlit}. Terms that require clarification
have been marked with a $\dagger$, and are explained at the end of this
table.}\vspace*{1mm}
\label{tab:evlitauthor}
\begin{tabular}{p{34mm}p{116mm}} 
\hline
    \textbf{Citation} & \textbf{Explanatory Virtues}  \\ \hline
    \citep{hempel1948studies} & Subsumption (Universal consistency) \\ \arrayrulecolor{lgray}\hline
    \citep{Ausubel1962} & Subsumption (Universal consistency) \\ \arrayrulecolor{lgray}\hline
    \citep{Kuhn1977} & Accuracy (Evidential accuracy), Internal consistency, Consistency
    with existing theories, Scope, Simplicity, Fruitfulness\\ \hline
    \citep{QuineUllian1978} & Conservatism$^\dagger$ (Universal consistency),
    Modesty, Simplicity, Generality (Scope), Refutability\\ \hline
    \citep{thagard1978best} & Consilience, Simplicity, Analogy  \\ \hline
    \citep{vanFraassen1980} & Epistemic -- Internal consistency, Consistency with
the facts, Empirical adequacy (Evidential accuracy), Empirical strength (Modesty);\\
 & Pragmatic -- Mathematical elegance, Simplicity, Scope, Unification,
Explanatory power (Conservativeness)\\ \hline
    \citep{Newton-Smith1981} & Observational nesting (Universal
consistency), Inter-theory support (Universal consistency), Internal
consistency, Fertility (Fruitfulness), Track record (Durability),
Smoothness, Compatibility with grounded metaphysical beliefs\\ \hline
    \citep{thagard1989coherence} & Explanatory coherence -- 
    Internal consistency, Universal consistency, Explanation (Causal adequacy),
    Data priority (Evidential accuracy), Analogy, Acceptability
(Completeness)\\ \hline
    \citep{lipton1991inference} & Mechanism (Causal adequacy), Precision
(Evidential accuracy), Unification$^\dagger$, Simplicity, Elegance \\ \hline
    \citep{Psillos2002} & Consilience$^\dagger$ (Universal consistency),
    Completeness, Importance, Parsimony (Simplicity), Unification,
    Precision$^\dagger$ (Causal adequacy)\\ \hline
    \citep{HitchcockWoodward2003} & Explanatory depth$^\dagger$ -- invariance
under changes in background knowledge\\ \hline
    \citep{schupbach2011logic} & Explanatory power (Conservativeness) \\ \hline
    \citep{mackonis2013inference} & Coherence$^\dagger$ (Universal consistency)
= Explanatory power -- derivable from Unification, Explanatory
depth (Causal adequacy), Simplicity\\ \hline
    \citep{McMullin2014} & Internal consistency, Contextual consistency
    (Universal consistency), Internal coherence$^\dagger$, Consilience, Optimality,
    Fertility (Fruitfulness), Durability\\ \hline
    \citep{vancleave2016} & Explanatoriness (Completeness -- explain
    all facts), Depth$^\dagger$, Power (Unification), Falsifiability (Refutability),
    Modesty$^\dagger$, Simplicity, Conservativeness \\ \hline
    \citep{RosalesMorton2021}\hspace*{-1mm} & Causal detail (Evidential
    accuracy), Causal generality (Causal adequacy), Unification\\
\arrayrulecolor{black}\hline
    \end{tabular}
\begin{minipage}{150mm}
\renewcommand{\baselinestretch}{0.78}\selectfont
{\footnotesize{$^\dagger$
\citeauthor{QuineUllian1978}'s (\citeyear{QuineUllian1978}) Conservatism
differs from Conservativeness~\citep{vancleave2016}; it requires that
a hypothesis conflict with as few previous beliefs as possible ---
a weaker version of Universal consistency.\\
\citeauthor{lipton1991inference}'s (\citeyear{lipton1991inference})
Unification is like \citeauthor{vancleave2016}'s (\citeyear{vancleave2016})
Depth, and differs from the Unification of \cite{vanFraassen1980},
\cite{Psillos2002}, \cite{mackonis2013inference} and \cite{RosalesMorton2021}.\\
\citeauthor{Psillos2002}'s (\citeyear{Psillos2002}) Consilience is similar to
Universal consistency (Fit with background knowledge), and differs from
\citeauthor{thagard1989coherence}'s (\citeyear{thagard1989coherence}); his
Precision differs from \citeauthor{lipton1991inference}'s
(\citeyear{lipton1991inference}).\\
\citeauthor{HitchcockWoodward2003}'s
(\citeyear{HitchcockWoodward2003}) Explanatory depth is less abstract than
\citeauthor{mackonis2013inference}'s (\citeyear{mackonis2013inference}).\\
\citeauthor{mackonis2013inference}'s (\citeyear{mackonis2013inference})
Coherence is similar to Universal consistency (Fit with background
knowledge), and equivalent to Explanatory power, which in turn differs from
\citeauthor{schupbach2011logic}'s (\citeyear{schupbach2011logic}),
and is derivable from other EVs.\\
\citeauthor{McMullin2014}'s (\citeyear{McMullin2014}) Internal coherence pertains
to the absence of \adhoc\ facts.\\
\citeauthor{vancleave2016}'s (\citeyear{vancleave2016}) Depth demands that an
explanation should not posit fundamentally new types of phenomena (raise more
questions than it answers); his Modesty demands that only claims relevant to
observed facts be mentioned, which is weaker requirement than 
that of \citeauthor{QuineUllian1978}'s (\citeyear{QuineUllian1978}) Modesty.
}
}
\end{minipage}
\renewcommand{\baselinestretch}{1}\selectfont
\end{table}

%% file: tables/ev-lit-tab.tex
\setlength{\tabcolsep}{4pt}
\begin{table}[ht]
    \centering
\renewcommand{\baselinestretch}{0.98}\selectfont
    \caption{Summary of Explanatory Virtues from psychology, philosophy and
cognitive science by type of virtue. The terms in one entry are similar
or equivalent; the author(s) listed at the beginning of the `Authors and Terms'
column use the term in the row heading.
\label{tab:evlit}}\vspace*{1mm}
    \begin{tabular}{p{34mm}p{116mm}} 
\hline
    \textbf{Explanatory Virtue} & {\textbf{Authors and Terms}}  \\ \hline
\multicolumn{2}{c}{\cellcolor{lgray}\textbf{Evidential virtues}} \\ \hline
Evidential accuracy  & Accuracy~\citep{Kuhn1977},
Empirical adequacy~\citep{vanFraassen1980}, Data
priority~\citep{thagard1989coherence}, Precision~\citep{lipton1991inference},
Causal detail~\citep{RosalesMorton2021}\\
Causal adequacy &
Explanation~\citep{thagard1989coherence}, Mechanism~\citep{lipton1991inference},
Precision~\citep{Psillos2002}, Explanatory depth~\citep{mackonis2013inference},
Causal generality~\citep{RosalesMorton2021}\\
Refutability & \citep{QuineUllian1978}, Falsifiability~\citep{vancleave2016}\\ 
Explanatory depth & \citep{HitchcockWoodward2003}\\
Optimality & \citep{McMullin2014}\\
\hline 
\multicolumn{2}{c}{\cellcolor{lgray}\textbf{Coherential virtues}}\\ \hline
Internal coherence  & \citep{McMullin2014}\\
Internal consistency  & \citep{Kuhn1977,vanFraassen1980,Newton-Smith1981,thagard1989coherence,McMullin2014}\\
Universal consistency  & Subsumption~\citep{hempel1948studies,Ausubel1962},
Consistency with existing theories~\citep{Kuhn1977},
Conservatism~\citep{QuineUllian1978}, Consistency with the facts~\citep{vanFraassen1980},
Observational nesting~\citep{Newton-Smith1981},
Inter-theory support~\citep{Newton-Smith1981},
Fit with background knowledge (Consilience, \cite{Psillos2002},
Coherence, \cite{mackonis2013inference}),
Contextual consistency~\citep{McMullin2014}\\
Analogy & \citep{thagard1978best,thagard1989coherence}\\
\hbox{Compatibility~with} metaphysical beliefs & \citep{Newton-Smith1981}\\
Conservativeness & \citep{vancleave2016},
Explanatory power~\citep{vanFraassen1980,schupbach2011logic}\\
Depth & \citep{vancleave2016}, Unification~\citep{lipton1991inference}\\
\hline 
\multicolumn{2}{c}{\cellcolor{lgray}\textbf{Coverage virtues}}\\ \hline
Completeness & \citep{Psillos2002}, Acceptability~\citep{thagard1989coherence},
Explanatoriness~\citep{vancleave2016}\\
Scope & \citep{Kuhn1977,vanFraassen1980}, Generality~\citep{QuineUllian1978}\\
Consilience & \citep{thagard1978best}\\
Unification &
\citep{vanFraassen1980,Psillos2002,mackonis2013inference,RosalesMorton2021},
Power~\citep{vancleave2016}\\
Importance & \citep{Psillos2002}\\
\hline 
\multicolumn{2}{c}{\cellcolor{lgray}\textbf{Aesthetic virtues}}\\ \hline
Simplicity &
\citep{Kuhn1977,thagard1978best,QuineUllian1978,vanFraassen1980,lipton1991inference,mackonis2013inference,vancleave2016},
Parsimony~\citep{Psillos2002}\\
Modesty & \citep{QuineUllian1978,vancleave2016}, Empirical
strength~\citep{vanFraassen1980}\\
\hline 
\multicolumn{2}{c}{\cellcolor{lgray}\textbf{Diachronic virtues}}\\ \hline
Durability & \citep{McMullin2014}, Track record~\citep{Newton-Smith1981}\\
Fruitfulness & \citep{Kuhn1977}, Fertility~\citep{Newton-Smith1981,McMullin2014}\\
Smoothness & \citep{Newton-Smith1981}\\
\hline 
     \end{tabular}
    \vspace*{5mm}
\end{table}